\documentclass[conference,9pt]{IEEEtran}
\usepackage{geometry}
\pdfoutput=1 
\IEEEoverridecommandlockouts
\usepackage{booktabs} 
\usepackage{algorithm}
\usepackage{algpseudocode}
\algtext*{EndWhile}
\algtext*{EndIf}
\algtext*{EndFor}
\algtext*{EndProcedure}
\usepackage{amsmath}
\usepackage{amssymb}
\usepackage{array}
\usepackage{bm}
\usepackage[bookmarks=false]{hyperref}
\usepackage[noadjust]{cite}
\usepackage{caption}
\usepackage{color}
\usepackage{comment}
\usepackage{dblfloatfix}
\usepackage[inline]{enumitem}
\usepackage{epsfig}
\usepackage{etoolbox}
\usepackage{fancybox}
\usepackage{float}
\usepackage{fixltx2e}
\usepackage{graphicx}
\usepackage{hyperref}
\usepackage{listings}
\usepackage{mathrsfs}
\usepackage{multirow}
\usepackage{multicol}
\usepackage{pifont}
\usepackage{placeins}
\usepackage{rotating}
\usepackage{setspace}
\usepackage{soul}
\usepackage[hang, tight]{subfigure}
\usepackage{threeparttable}
\usepackage{times}
\usepackage{url}
\usepackage{upgreek}
\usepackage{verbatim}
\usepackage{wrapfig}
\usepackage[usenames,dvipsnames]{xcolor}
\usepackage{authblk}
\usepackage[]{siunitx}
\usepackage[super]{nth}
\usepackage{tabularx}
\usepackage{graphicx}
\usepackage{supertabular}
\usepackage{tikz}
\usepackage{bbm}
\newbool{inccomment}
\setbool{inccomment}{true}
\newcommand{\XX}[1]{\ifbool{inccomment}{{\color{magenta} #1}}{}}
\newcommand{\CT}[1]{\ifbool{inccomment}{{\color{magenta}CT\@: #1}}{}}
\newcommand{\NT}[1]{\ifbool{inccomment}{{\color{blue}NT\@: #1}}{}}
\newcommand{\TD}[1]{\ifbool{inccomment}{{\color{orange}#1}}{}}
\newcommand{\FN}[1]{\ifbool{inccomment}{{\color{OliveGreen}#1}}{}}
\newcommand{\GR}[1]{\ifbool{inccomment}{{\color{Tan}#1}}{}}
\newcommand{\LD}{\ifbool{inccomment}{{\color{magenta}\\============================================\\}}}
\newcommand{\RF}{\ifbool{inccomment}{{\color{green}~[R]}}}

\newcommand{\roma}[1]{\uppercase\expandafter{\romannumeral #1\relax}}

\graphicspath{{./_fig/}}
\DeclareMathOperator*{\argmin}{arg\,min}

\usepackage{amsthm}
\newtheorem{thm}{Theorem}[section] 

\newcommand{\thistheoremname}{}
\newtheorem{genericthm}[thm]{\thistheoremname}

\newtheorem*{genericthm*}{\thistheoremname}
\newenvironment{namedthm*}[1]
  {\renewcommand{\thistheoremname}{#1}%
   \begin{genericthm*}}
  {\end{genericthm*}}

\usepackage{mathtools}
\DeclarePairedDelimiter{\ceil}{\lceil}{\rceil}
\DeclarePairedDelimiter{\floor}{\lfloor}{\rfloor}

\graphicspath{{./_fig/}}
\newcommand{\fixme}[1]{\textcolor{black}{#1}}
\usepackage{amsthm}

\allowdisplaybreaks
\makeatletter
\def\BState{\State\hskip-\ALG@thistlm}
\makeatother
\begin{document}

\title{PowerNet: Transferable Dynamic IR Drop Estimation via Maximum Convolutional Neural Network
}

\author[]{ \fontsize{12}{12}\selectfont Zhiyao Xie$^1$, Haoxing Ren$^2$, Brucek Khailany$^2$, Ye Sheng$^2$, Santosh Santosh$^2$, Jiang Hu$^3$, Yiran Chen$^1$ \vspace{-3pt}}

\affil[]{\fontsize{10}{10}\selectfont $^1$Duke University, $^2$Nvidia Corporation, $^3$Texas A\&M University \vspace{-3pt}}

\affil[]{\{zhiyao.xie, yiran.chen\}@duke.edu, 
         \{haoxingr, bkhailany, sye, santosha\}@nvidia.com,
         jianghu@tamu.edu\vspace{-7pt}}

\makeatletter
\def\ps@IEEEtitlepagestyle{%
  \def\@oddfoot{\mycopyrightnotice}%
  \def\@evenfoot{}%
}

\makeatother
\def\mycopyrightnotice{%
  \begin{minipage}{\textwidth}
    \footnotesize
    978-1-7281-4123-7/20/\$31.00~\copyright~2020 IEEE \hfill\\~\\
  \end{minipage}
  \gdef\mycopyrightnotice{}
}


\maketitle

\begin{abstract}
IR drop is a fundamental constraint required by almost all chip designs. However, its evaluation usually takes a long time that hinders mitigation techniques for fixing its violations. 
In this work, we develop a fast dynamic IR drop estimation technique, named PowerNet, based on a convolutional neural network (CNN). It can handle both vector-based and vectorless IR analyses. Moreover, the proposed CNN model is general and transferable to different designs. This is in contrast to most existing machine learning (ML) approaches, where a model is applicable only to a specific design.
Experimental results show that PowerNet outperforms the latest ML method by 9\% in accuracy for the challenging case of vectorless IR drop and achieves a $30\times$ speedup compared to an accurate IR drop commercial tool. Further, a mitigation tool guided by PowerNet reduces IR drop hotspots by 26\% and 31\% on two industrial designs, respectively, with very limited modification on their power grids.
\end{abstract}

\section{Introduction}

Dynamic IR drop is the deviation of the power supply level from its specification caused by localized power demand and switching patterns. It must be restricted in order for a circuit to meet its timing target and function
properly. As such, it is vitally important to verify if IR drop
satisfies design constraints and identify constraint violation regions, a.k.a. hotspots. As chip complexity continues to grow, IR drop evaluation becomes increasingly challenging.


In industrial designs, dynamic IR drop estimation is often obtained by running simulation-based commercial tools, which are known to be accurate but very time-consuming. 
%
Machine learning (ML)-based approaches have been explored in an effort to achieve faster estimation. 
Many of these previous works are summarized in Table \ref{tbl:t_prev}. These works learn to predict dynamic IR drop of each cell through features such as cell positions, timing windows, path resistance, etc. with supervised machine learning techniques. 


A major weakness shared by most of the previous works is that
they are not ``design independent'', i.e., {\em transferable} to new designs that are not seen in its training dataset. 
In other words, most of these previous works need to train a new model for each distinct design.
Some work~\cite{yamato2012fast} even dedicates one model for every single cell. Training a new model with new labels entails a long simulation and training time, which defeats the original purpose of fast estimation. The only exception is \cite{dhotre2017identification}, which is based on unsupervised learning and does not learn any previous knowledge. 


In addition, most previous ML approaches to IR drop estimation only focus on vector-based analysis, ignoring vectorless IR drop. For dynamic IR drop, the peak IR drop in the design can be analyzed either using vectorless analysis or vector-based analysis using simulation patterns from value change dump (VCD) files. Vectorless IR drop analysis is highly desirable for IR mitigation during physical design for two main reasons. Firstly, for a large chiplet, vector-based IR drop analysis requires
a huge number of simulation patterns to cover most regions and thus can be unbearably slow. 
Secondly, designers are unable to obtain accurate power simulation patterns early in the design process. For large industrial designs, multiple teams work on different RTL units in parallel and the overall simulation patterns change throughout the design process. 
Vectorless IR drop provides a faster and earlier estimation in this case, however, accurate estimation is more difficult than vector-based due to the increased diversity in switching activity distribution. We will demonstrate the accuracy difference between vector-based and vectorless IR drop analysis in Section \ref{vector}. 

\begin{table}[!tb]
  \centering
  \caption{Comparison Among Different Works}
  \vspace{-1mm}
  \label{tbl:t_prev}
  \resizebox{0.95\linewidth}{!}{
  \begin{tabular}{l | c c}
 	\hline
	ML Methods & Model & Design Independent   \\  
	\hline
	\cite{yamato2012fast} (ITC 12) &    Linear Regression  &  No \\
	\cite{ye2014chip} (VTS 14)  &    SVM  &   No \\
	\cite{dhotre2017identification} (ATS 17) &  Clustering  & Unsupervised  \\
	\cite{lin2018ir} (VTS 18) &  ANN   &   No \\
	\cite{fang2018machine} (ICCAD 18) &  XGBoost   &   No \\
	\hline
	PowerNet & Max-CNN & Yes \\ 
	\hline
	
  \end{tabular}
  }
\end{table}

Our CNN-based method PowerNet provides a transferable ML model for both vectorless and vector-based IR drop estimations. We put more emphasis on vectorless estimation in our experiments, considering its higher difficulty and usability. PowerNet addresses these challenges by its innovative preprocessed features and CNN architecture. In previous works \cite{fang2018machine}, the design dependent features such as coordinates and timing information of each cell are directly fed into the ML model. Since locations and timing do not directly cause IR drop, directly fitting a model based on these features would likely introduce the {\em overfitting} problem, making the model inaccurate on unseen designs. 
Instead, design-dependent information should be preprocessed to correlate with IR drop before feeding to ML models. It is known that IR drop directly correlates with cell power consumption. Therefore, PowerNet carefully incorporates these design-dependent features into power maps during preprocessing. 
It also utilizes an innovative CNN architecture to capture maximum transient IR drop. The main contributions of our work include:

\begin{itemize}
\item We propose PowerNet, an innovative CNN method targeting both vectorless and vector-based IR drop estimation. It is the first method that claims to perform design-independent fast IR drop estimations. 
\item For experiments on both vectorless and vector-based IR drop estimations, PowerNet outperforms all other ML methods on every tested industrial design. Especially for vectorless prediction, PowerNet gives a 9\% higher accuracy.
\item PowerNet is 
30$\times$ faster than an accurate simulation-based commercial IR drop analysis tool.
\item An IR drop mitigation tool guided by PowerNet reduces IR drop hotspots by 26\% and 31\% on two new industrial designs with very limited modifications to the power grid.
\item We provide a detailed analysis on PowerNet's mechanism by showing two representative examples.
\end{itemize}

\section{Problem Formulation}

This work aims at detecting locations of IR drop hotspots. Hotspots are regions where IR drop is greater than a specified threshold. To estimate IR drop, every design is tessellated into an array of tiles, each of which is an $l\times l$ square. The tile size $l$ controls the granularity of our solution. In this way, a design with the size of $W \times H$ is represented as a $w \times h$ matrix, where $w = W/l$ and $h = H/l$. The IR drop at each tile is the mean value of IR drop of all cells within it. Then IR drop for the whole design is $IR \in \mathbf{R} ^{w \times h}$. \fixme{The ground-truth $IR$ is also referred to as label in this paper}. As for input features, different types of power dissipation values are calculated for each tile. We refer to each $w \times h$ power matrix as a power map. Essentially, power maps are the distribution of power density. PowerNet $F$ tries to give the closest estimation $F^*$ on $IR$ based on all $G$ different power maps $\{P_{map1} \,... \,P_{mapG}\}$. 

\begin{gather*} 
F:  \{P_{map1}\in \mathbf{R} ^{w \times h} ... \, P_{mapG} \in \mathbf{R} ^{w \times h}\}  \rightarrow \mathbf{R} ^{w \times h} \\ 
F^* = \argmin_{F} Loss(F(\{P_{map1} ... \, P_{mapG}\}), IR).
\end{gather*}

\section{Algorithm}

\subsection{Feature Extraction}

According to Ohm's Law, excessive IR drop can be caused by either high current or high resistance. As is typical in state-of-the-art VLSI design, we assume a uniform power grid in the power delivery network (PDN), which means the resistance distribution across a whole design is also rather uniform. Thus in PowerNet, we choose not to spend extra time calculating resistance for each cell. For designs with a non-uniform PDN, each cell's power value can be scaled by its resistance. The influence of resistance is further elaborated in Section \ref{resistance}. When resistance is considered consistent, current becomes the only key issue in IR drop estimation. Since local power consumption is proportional to local current, PowerNet utilizes cell power as its input features.

For each cell $c$, \fixme{we do not exhaust all possible features that seem to be relevant, which make the model too complex and overfit. Instead, we select features that prove to provide essential information for IR drop estimation. Hard macros are not included. Below} are the details of all features and the labels extracted from them:
\begin{itemize}
\item \textbf{Power}: Three types of power values are extracted.
    \begin{itemize}
    \item Internal power ($p_i$)
    \item Switching power ($p_s$)
    \item Leakage power ($p_l$)
    \end{itemize}
\item \textbf{Signal arrival time}: The minimum and maximum signal arrival times to the cell within a clock cycle.
    \begin{itemize}
    \item Min arrival time ($t_{min}$)
    \item Max arrival time ($t_{max}$)
    \end{itemize}
\item \textbf{Coordinates}: The cell location after placement.
    \begin{itemize}
    \item Min and max x axis ($x_{min}, x_{max}$)
    \item Min and max y axis ($y_{min}, y_{max}$)
    \end{itemize}
\item \textbf{Toggle rate}: Describes how often output changes with regard to a given clock input.
    \begin{itemize}
    \item Rate ($r_{tog}$)
    \end{itemize}
\item \textbf{IR drop}: The difference between the nominal supply voltage and the actual voltage arrived at each cell. ($ir$)
\end{itemize}

All of the above features are scalar values. For these power types, internal power $p_i$ means power dissipated by capacitance internal to each cell; switching power $p_s$ is power dissipated by the load capacitance at the output of the cell; leakage power $p_l$, which is relatively small in the experiment, is consumed by unintended leakage that does not contribute to function. Based on these basic power types, we can generate more power information for each cell: 
\begin{gather*} 
p_{sca} = (p_i + p_s)*r_{tog} + p_l \\ 
p_{all} = p_i + p_s + p_l
\end{gather*}
Both $p_{sca}$ and $p_{all}$ reflect the overall power dissipated by cells, but $p_{sca}$ scales the overall power by toggle rate of each cell. PowerNet learns to combine the total power from these different sources of power dissipation.

\subsection{Preprocessing by Decomposition}

\begin{figure}[!b]
  \begin{minipage}[b]{3.5in}
  \centering
    \includegraphics[width=3.2in]{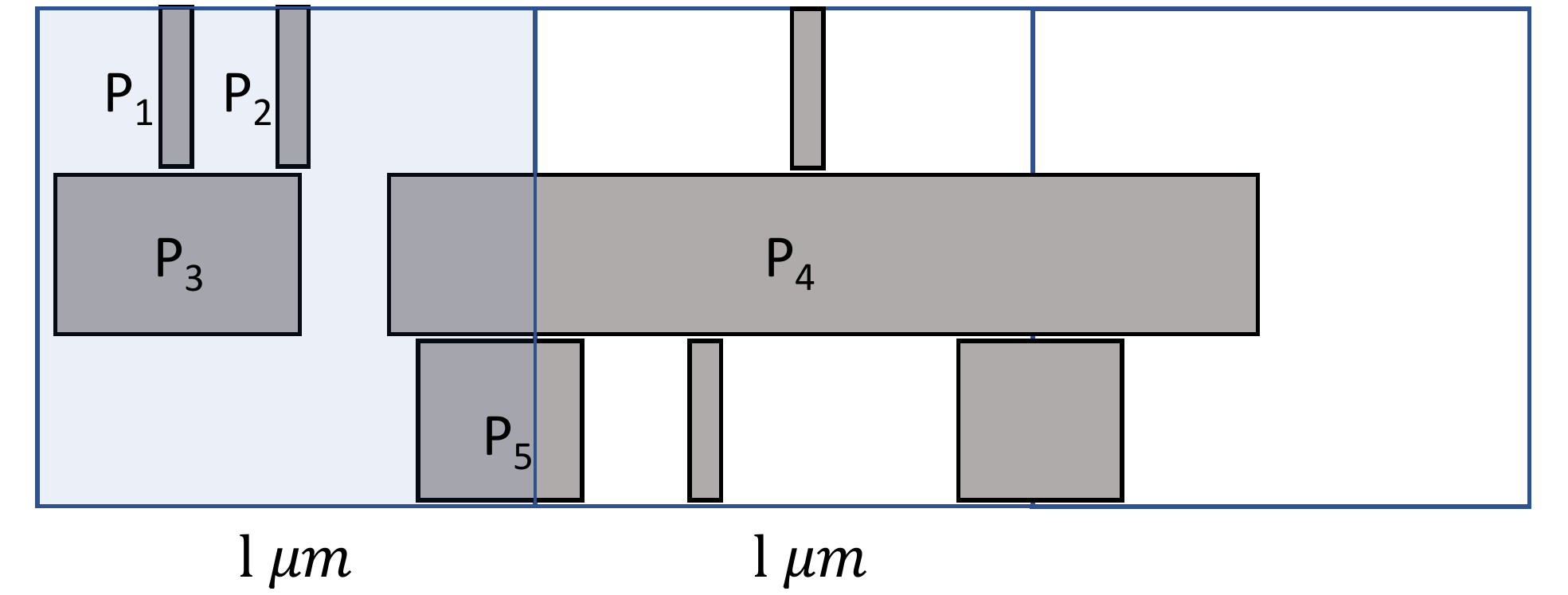}
    \vspace{-2mm}
  \caption{Space decomposition.}
  \label{spaceDecomp}
  \end{minipage}
\end{figure}

After power is extracted, the IR drop seen at each cell is not just simply proportional to its own cell power but also depends on its neighborhood due to both spatial and temporal current distributions. Spatially, local current is proportional to the sum of power demand of all cells in a local region. Hence, the power of neighboring cells also contributes to IR drop of the analyzed cell. We amortize cell power into grid tiles by a space decomposition. This also motivates us to adopt a CNN model in PowerNet, which is inherently designed for learning scalable two-dimensional patterns. Even when considering spatial information, a region with high overall power demand may still not be IR drop hotspot.  This case arises when cells in the region do not switch at the same time. Such asynchronous switching disperses voltage drop into a larger timing window. As a result, maximum dynamic IR drop, i.e. the highest-transient voltage drop, can still be low. PowerNet measures such influence by time decomposition during preprocessing.

Algorithm \ref{alg1} shows our preprocessing method. It generates power maps based on cell information. For each design, two types of power maps are generated. The first type includes $\{P_{i}$, $P_{s}$, $P_{sca}$, $P_{all}\}$. They only go through spatial decomposition and do not carry timing information. The second type $\{P_{t}[\,j] \in \mathbb{R} ^ {w\times h} \mid j \in [1, N]\}$ goes through both a space decomposition and a time decomposition.

\begin{algorithm}[!t]
\caption{Preprocessing by Decomposition}
\label{alg1}
\begin{flushleft}
  \textbf{Input}: Features \{$p_i$, $p_s$, $p_l$, $t_{min}$, $t_{max}$, $x_{min}$, $x_{max}$, $y_{min}$, $y_{max}$, $r_{tog}$\} for every cell $c$. Design weight $W$, height $H$, cell number $C$ and clock cycle $T$. Tile size $l$ and time window $t$. \\
  \textbf{Preprocess}:
\end{flushleft}
\begin{algorithmic}[1]
\State $w = W/l$, $h = H/l$, $N = T/t$ 
\State Set $P_{i}$, $P_{s}$, $P_{sca}$, $P_{all}$ $\in \{0\} ^ {w\times h}$
\State Set $\{P_{t}[\,j] \in \{0\} ^ {w\times h} \mid j \in [1, N]\}$
\For{each cell $c \in [1,C]$}
\State $p_{sca} = (p_i + p_s) * r_{tog} + p_l $
\State $p_{all} = p_i + p_s + p_l$
\State $x_{n} = \floor{(x_{min}/l)}$, $x_{x} = \ceil{(x_{max}/l)}$ \label{line:s_b}
\State $y_{n} = \floor{(y_{min}/l)}$, $y_{x} = \ceil{(y_{max}/l)}$
\State $s = (x_{x} - x_{n}) * (y_{x} - y_{n})$
\State Set mask $M \in \{0\} ^ {w\times h}$, $M[x_n:x_x][y_n:y_x] = 1$
\State $P_{i}$ += $M* p_{i}/s$
\State $P_{s}$ += $M* p_{s}/s$
\State $P_{sca}$ += $M* p_{sca}/s$
\State $P_{all}$ += $M* p_{all}/s$  \label{line:s_e}
\For{each int $j \in [1,N]$}        \label{line:t_b}
\If {$t_{min} < j*t$ and $t_{max} > j*t$}
\State $P_{t}[\,j]$ += $M* p_{sca}/s$   \label{line:t_e}
\EndIf
\EndFor
\EndFor
\end{algorithmic} 
\begin{flushleft}
  \textbf{Output}: Time-decomposed $\{P_{t}[\,j] \in \mathbb{R} ^ {w\times h} \mid j \in [1, N]\}$, \\
  \hspace{12.5mm} Power map $P_{i}$, $P_{s}$, $P_{sca}$, $P_{all}$ $\in \mathbb{R} ^ {w\times h}$
\end{flushleft}
\end{algorithm}

\begin{figure}[!b]
  \begin{minipage}[t]{3.5in}
  \centering
    \includegraphics[width=3in]{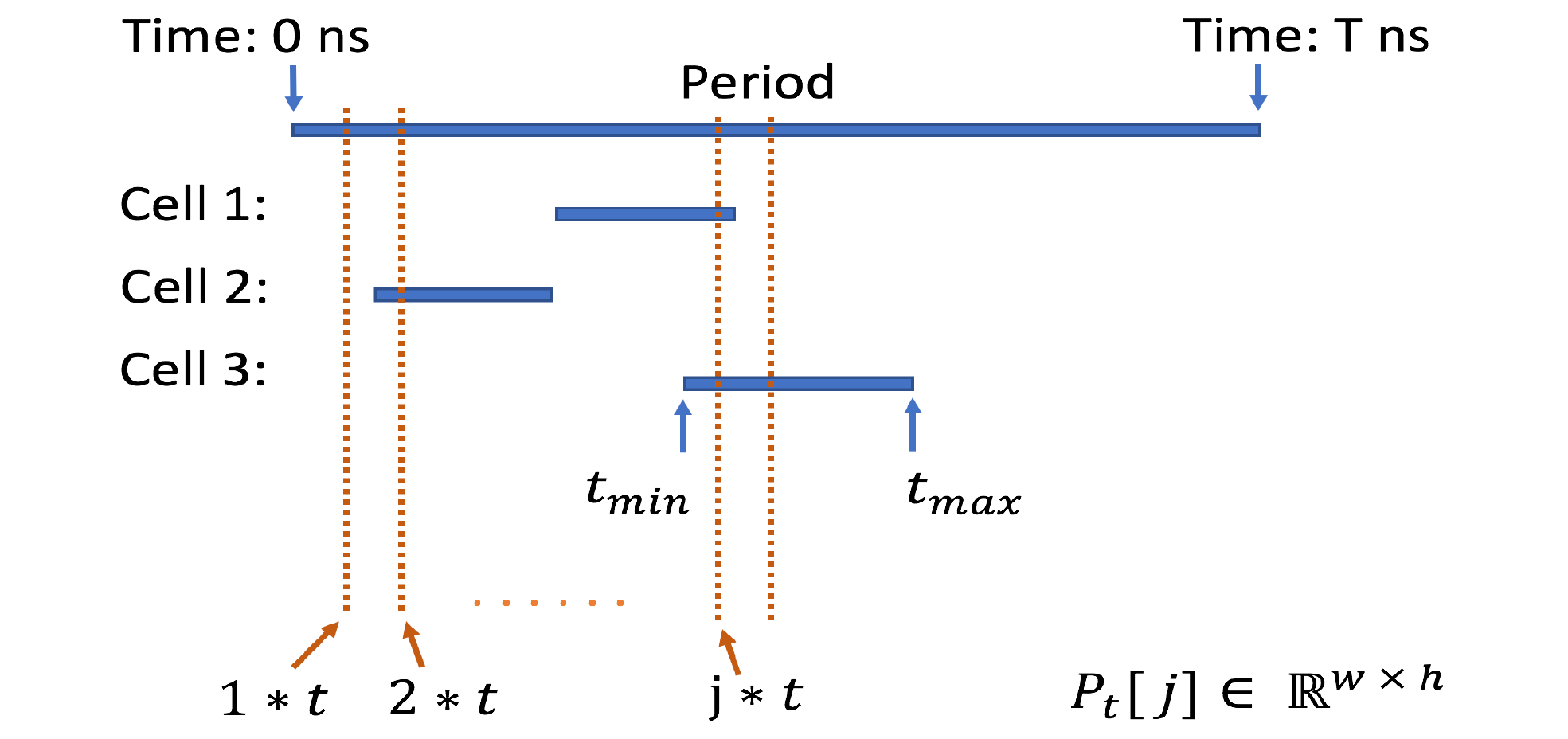}
    \vspace{-2mm}
  \caption{Time decomposition.}
  \label{timeDecomp}
  \end{minipage}
\end{figure}

As illustrated in Figure \ref{spaceDecomp}, space decomposition (Lines \ref{line:s_b} to \ref{line:s_e}) amortizes cell power into any grid tiles occupied by the cell. Assume the regular squares are grid tiles and grey rectangles are cells. $P_1$ to $P_5$ are cell power. For the leftmost highlighted tile, its power equals $P_1 + P_2 + P_3 + P_4/3 + P_5/2$. The long cell with power $P_4$ only contributes  one-third of its power to that of the highlighted tile, because altogether it overlaps with three tiles. Similarly, in line \ref{line:s_b} to \ref{line:s_e}, each cell contributes $p/s$, where $s$ is the number of overlapping tiles.

Lines \ref{line:t_b} to \ref{line:t_e} perform time decomposition. Every power map $P_t[\,j]$ corresponds to one time instant $j*t$. For each cell at $j*t$, it contributes power to a corresponding power map $P_t[\,j]$ only when $j*t$ falls between its signal arriving time $[t_{min}, t_{max}]$. In other words, only cells that can possibly switch at that instant are considered. Figure \ref{timeDecomp} demonstrates the mechanism. Vertical dashed lines are measured instants $1*t$ to $j*t$, and horizontal bars are the signal arrival time intervals of cells. In this example, only cells 1 and 3 will be counted for $P_t[\,j]$ and no cells are counted for $P_t[1]$. 

\begin{figure}[!t]
  \begin{minipage}[t]{3.5in}
  \centering
    \includegraphics[width=3.3in]{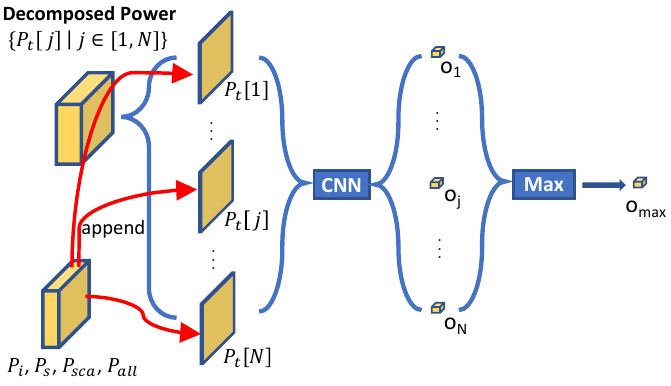}
    \vspace{-2mm}
  \caption{PowerNet structure.}
  \label{PowerNet}
  \end{minipage}
\end{figure}

\subsection{PowerNet Model}

\begin{algorithm}[!b]
\caption{PowerNet: Algorithm of Maximum CNN}
\label{alg2}
\begin{flushleft}
  \textbf{Input}: IR drop label $IR \in \mathbb{R} ^ {w\times h}$, Power $P_{i}$, $P_{s}$, $P_{sca}$, $P_{all}$ $\in \mathbb{R} ^ {w\times h}$, Time-decomposed $\{P_{t}[\,j] \in \mathbb{R} ^ {w\times h} \mid j \in [1, N]\}$, Input window size $k =2*k_h+1$, $k_h$ means half size
\end{flushleft}
\begin{flushleft}
\textbf{Training}:
\end{flushleft}
\begin{algorithmic}[1]
\Function{getInput}{$\,j, x, y$}
    \State Stack features $I = \{ P_{i}$, $P_{s}$, $P_{sca}$, $P_{all}$, $P_{t}[j] \} \in \mathbb{R} ^ {w\times h \times 5}$\label{line:l0}
    \State $I_{x,y} = I[x-k_h:x+k_h+1][y-k_h:y+k_h+1] \in \mathbb{R} ^ {k\times k \times 5}$ 
    \State return $I_{x, y}$
\EndFunction
\State
\State Initiate CNN model $f$: $\mathbb{R} ^ {k\times k \times 5} \rightarrow \mathbb{R}$, Loss function $J$
\For{$epoch \in [1,N_{epoch}]$}
\For{$x \in $ shuffle $([1, w])$, $y \in $ shuffle $([1, h])$}
\State $o_{max} = 0$
\For{each $j \in [1,N]$}                              \label{line:l1}
\State $I_{x, y} = $ \textproc{getInput} $ (\,j, x, y)$ \label{line:l2}
\State $o_{j} = f(I_{x, y})$
\If{$o_{max} < o_{j}$}
\State $o_{max} = o_{j}$
\EndIf
\EndFor
\State *Gradient Descent $f$ -= $\nabla J(o_{max}, IR[x, y])$
\EndFor
\EndFor
\end{algorithmic}
\begin{flushleft}
  \textbf{Output}: Trained CNN model $f$: $\mathbb{R} ^ {k\times k \times 5} \rightarrow \mathbb{R}$
\end{flushleft}
\end{algorithm}

Algorithm \ref{alg2} shows how PowerNet $F$ handles power maps with its CNN model $f$. For each training epoch, it iterates through every tile $(x, y)$ in training designs. For every tile, it crops $k\times k$ input windows surrounding it from all relevant $w \times h$ power maps by the function \textproc{getInput}.

As shown in Lines \ref{line:l1} to \ref{line:l2} and Figure \ref{PowerNet}, for all $N$ time-decomposed power maps $\{P_{t}[\,j] \in \mathbb{R} ^ {w\times h} \mid j \in [1, N]\}$, they are processed separately by the same CNN model, together with all other common power maps $P_{i}$, $P_{s}$, $P_{sca}$, $P_{all}$. Hence, the input to the CNN is $\{P_{i}$, $P_{s}$, $P_{sca}$, $P_{all}$, $P_{t}[\,j] \}$ in Line \ref{line:l0}. It results in a total of $N$ CNN outputs $\{o_j \mid j \in [1, N]\}$. Then, the maximum output $o_{max} = Max(\{o_j \mid j \in [1, N]\})$ is the prediction result for the analyzed tile. This maximum structure highlights the only instant that leads to the peak IR drop. It guides CNN $f$ to learn such a pattern.

\begin{figure}[!t]
  \begin{minipage}[t]{3.5in}
  \centering
    \includegraphics[width=3.3in]{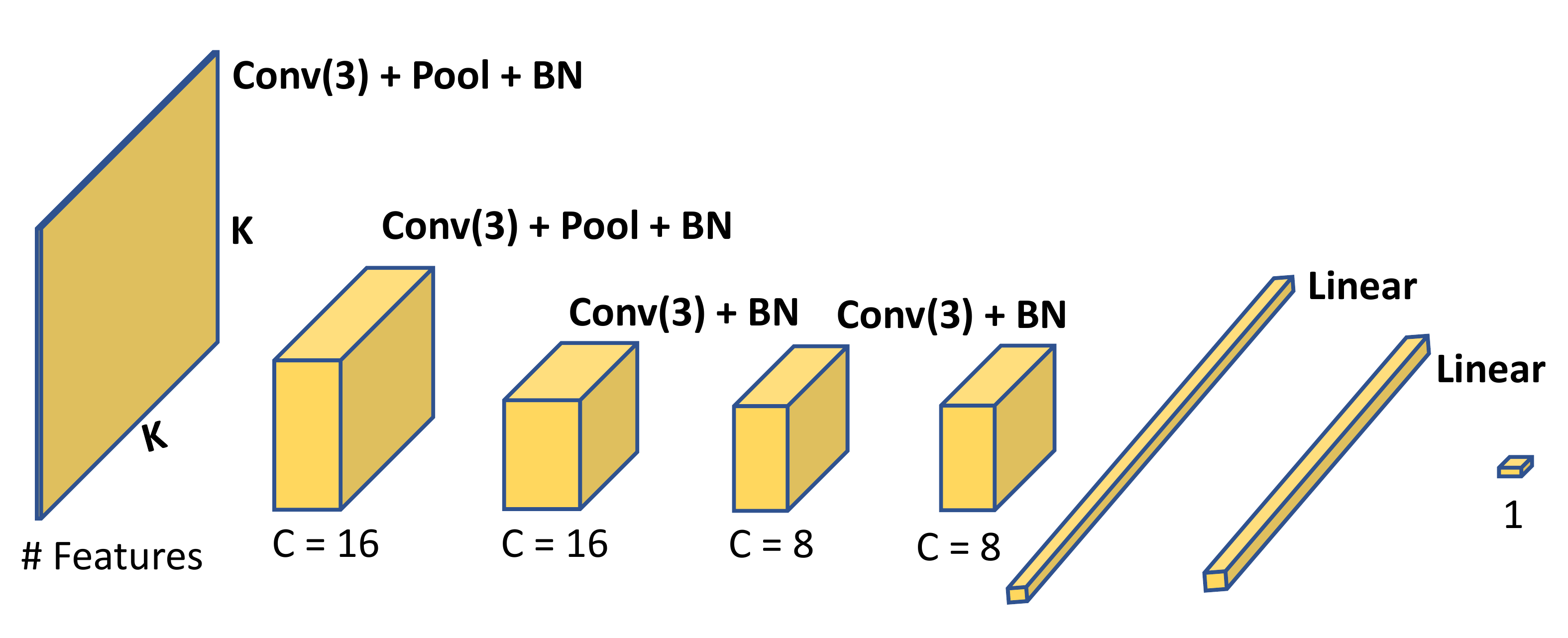}
  \caption{CNN structure.}
  \label{CNN}
  \end{minipage}
\end{figure}

Details of the CNN model $f$ in PowerNet is shown in Figure \ref{CNN}. There are four convolutional layers, two pooling layers and two fully connected layers. Size of convolution kernels is given in parentheses. The $C$ under each tensor gives the number of kernels defined in each convolutional layer. \fixme{This CNN structure and hyperparameters like $N$, $k$ are tuned based on the performance during cross-validation. Choosing larger input $k$, more layers or kernels turns out to reduce model generalization and slow down the prediction, while a simpler structure underfits the data.} Batch normalization (BN) \cite{batch_norm} is applied to accelerate model convergence. Adam method \cite{Adam} is used for optimization.
We adopt the mean absolute error between prediction and label (L1 loss) as loss function.
\section{Experimental Results}
 
\subsection{Experiment Setup}

\begin{table}[!h]
\vspace{-2mm}
  \centering
  \caption{Designs Used in Experiment}
  \vspace{-2mm}
  \label{tbl:t1}
  \resizebox{\linewidth}{!}{%
  \begin{tabular}{l | c c c c | c c}
 	\hline
	Design   & D1 & D2 & D3 & D4 & MD1 & MD2   \\
	\hline
    \# cells (million) &  1.7 &  0.81  & 2.0  & 1.9 & 1.7 & 2.4 \\
    Hotspot	Portion    &  5.6\%  & 7.7\%  & 3.1\%  & 3.1\% & 0.65\% & 0.50\%  \\
	\hline
  \end{tabular}
  }
\end{table}

In our experiment, we use six industrial designs in a sub-10nm technology node (Table \ref{tbl:t1}) with an IR drop hotspot threshold of $56mV$, 6\% of the supply voltage ($0.94 V$). Features and IR drop labels are extracted after clock tree synthesis (CTS). Though tested at the CTS stage, PowerNet can also be applied to other stages. We perform vectorless analysis and use results from a commercial IR drop analysis tool as labels. We train the models and measure their accuracy on D1 to D4, then mitigate the IR drop of MD1 and MD2 with the estimation from PowerNet.
When testing estimation accuracy on D1 to D4, the ML model is trained only on data from the other three designs. It ensures that the tested design is totally {\em unseen} to the corresponding model, which eliminates the possibility of information leakage between the testing and training datasets.

We implemented CNN and \fixme{tree-based} XGBoost models similar to \cite{fang2018machine} as a comparison with PowerNet. Similar to \cite{fang2018machine}, two types of features are extracted, named  cell features and  map features. \fixme{Cell features are one-dimensional and map features are two-dimensional.}
For each cell $c$, cell features include \fixme{its} signal arrival time, coordinates, capacitance, unscaled overall cell power $p_{all}$, toggle rate $r_{tog}$ and cell type. The current of each cell is not included because it is not available in our design flow. Since voltages at different regions are all close to the supply voltage, current can be viewed as proportional to power. Then, local maps of both unscaled overall power $p_{all}$ and $r_{tog}$ around it are constructed as its map feature. Notice that compared with PowerNet, only one type of power $p_{all}$ is used. For the \fixme{tree-based} XGBoost model, all map features and cell features are directly used as model input. To fit into XGBoost, the two-dimensional map features are flattened into one dimension. For the CNN model, map features firstly go through three convolutional layers, each with 25, 25 and 50 filters. Then, the output of convolutional layers together with all cell features are fed into three fully connected layers, each with 512 neurons. A 0.4 dropout rate~\cite{srivastava2014dropout} is applied. \fixme{Other hyper-parameters like optimizer or learning rate of baselines are carefully tuned for their best performance. They are trained and tested on the same designs as PowerNet for model comparison.} 
 
All algorithms are implemented in Python. CNN-related models are built on PyTorch 1.0 \cite{paszke2017automatic}. For PowerNet, we set tile size $l=1 \,\si{\micro\meter}$, number of measured instants $N=50$, and an input window size $k=31$ in the experiment. Both training and testing are implemented on an 8-core CPU machine with 100 GB memory and one NVIDIA Tesla V100 GPU.

 \subsection{Result Measured in ROC Curve}
 
 \begin{figure}[!t]
  \centering
  \begin{minipage}[t]{3.5in}
  \centering
    \includegraphics[width=3in]{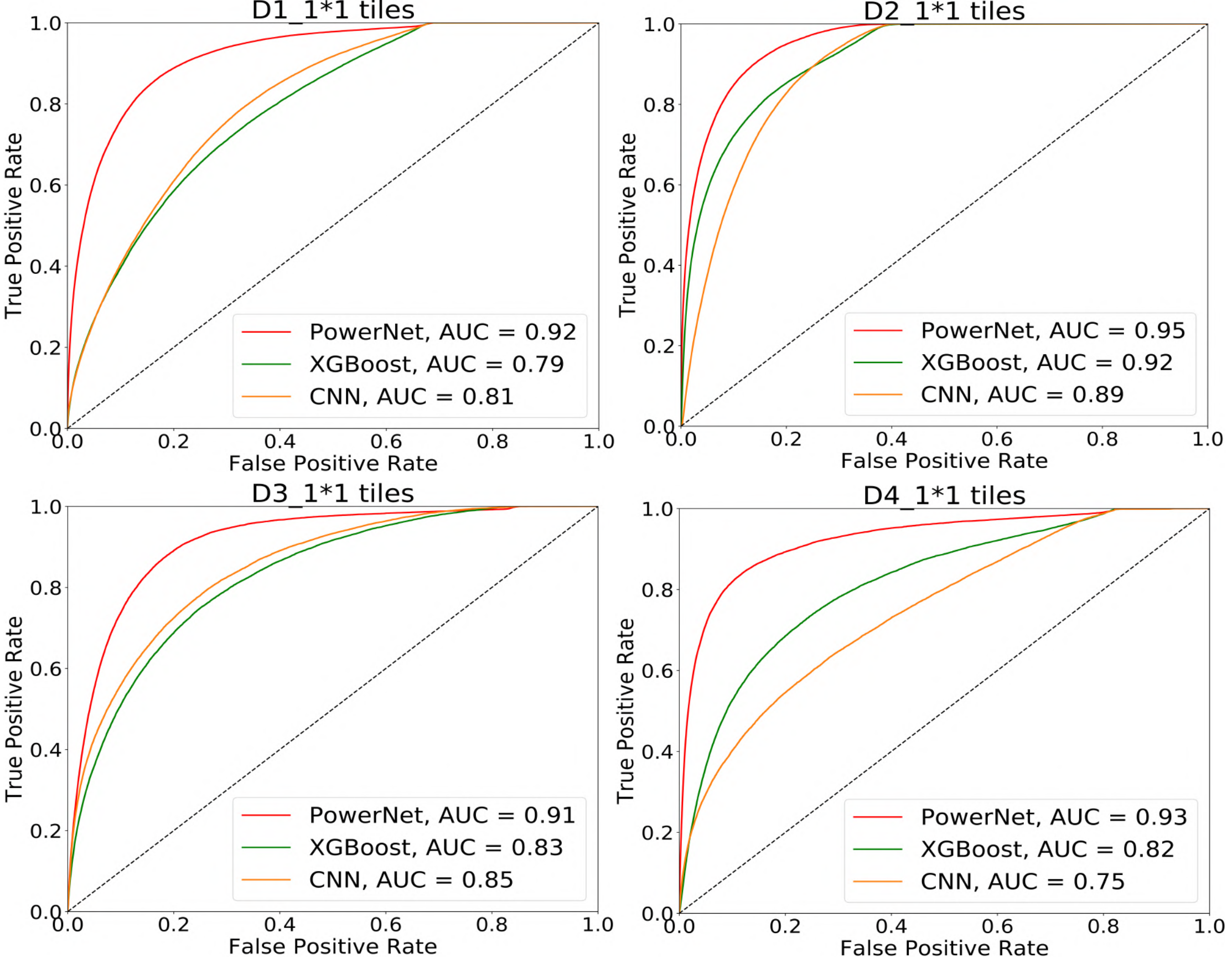}
    \vspace{-1mm}
  \caption{Comparison of methods by ROC curve. Measured in $1\times 1$ tiles granularity.}
  \label{ROC_1}
  \vspace{1mm}
    \includegraphics[width=3in]{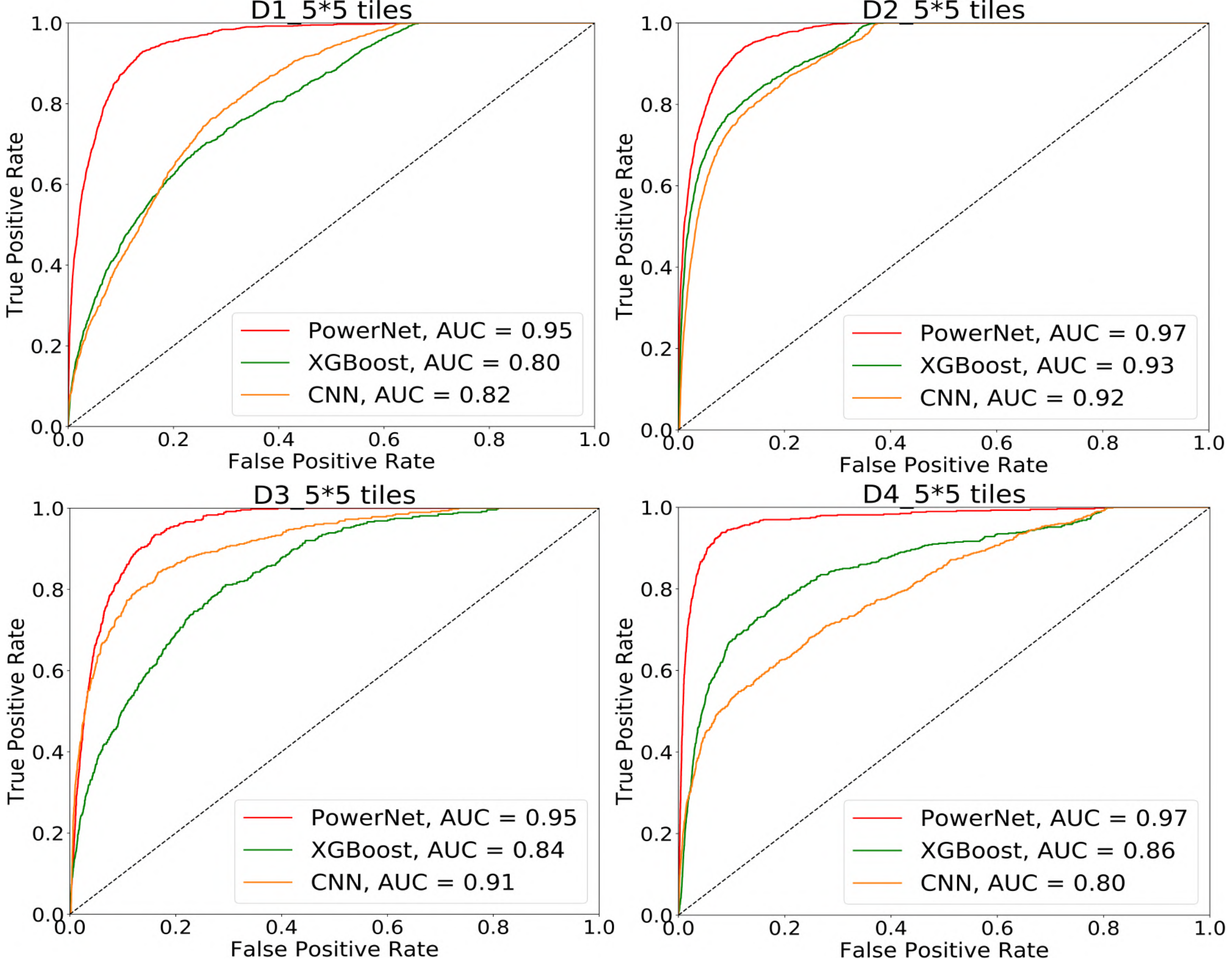}
    \vspace{-1mm}
  \caption{Comparison of methods by ROC curve. Measured in $5\times 5$ tiles granularity.}
  \label{ROC_5}
  \end{minipage}
\end{figure}

Figures \ref{ROC_1} and \ref{ROC_5} show the performance on different designs when measured in $1\times 1$ tiles and $5\times 5$ tiles, respectively. Measurement in $5\times 5$ tiles means tessellating both predictions and labels with a larger tile, whose size is $5l \times 5l$. In this case $IR \in \mathbf{R}^{\frac{w}{5}\times \frac{h}{5}}$. 
Sometimes when designers fix IR drop by performing power grid (PG) enhancements, hotspots shown in $5l\times5l$ tile already provide sufficient information.
Accuracy is measured by the area under the ROC curve (AUC $\in [0, 1]$). A larger area means a better accuracy \fixme{in hotspot identification}. PowerNet achieves AUC higher than 0.9 and 0.95 for all designs for the aforementioned two granularity's, respectively. On average, for $1\times1$ tiles, the AUC for CNN, XGBoost and PowerNet are around 0.83, 0.84 and 0.93. For $5\times5$ tiles, their AUC are around 0.86, 0.86 and 0.96. PowerNet's improvement in accuracy is $9\%$. 

\begin{figure}[!t]
  \centering
  \begin{minipage}[t]{3.5in}
    \includegraphics[width=3.5in]{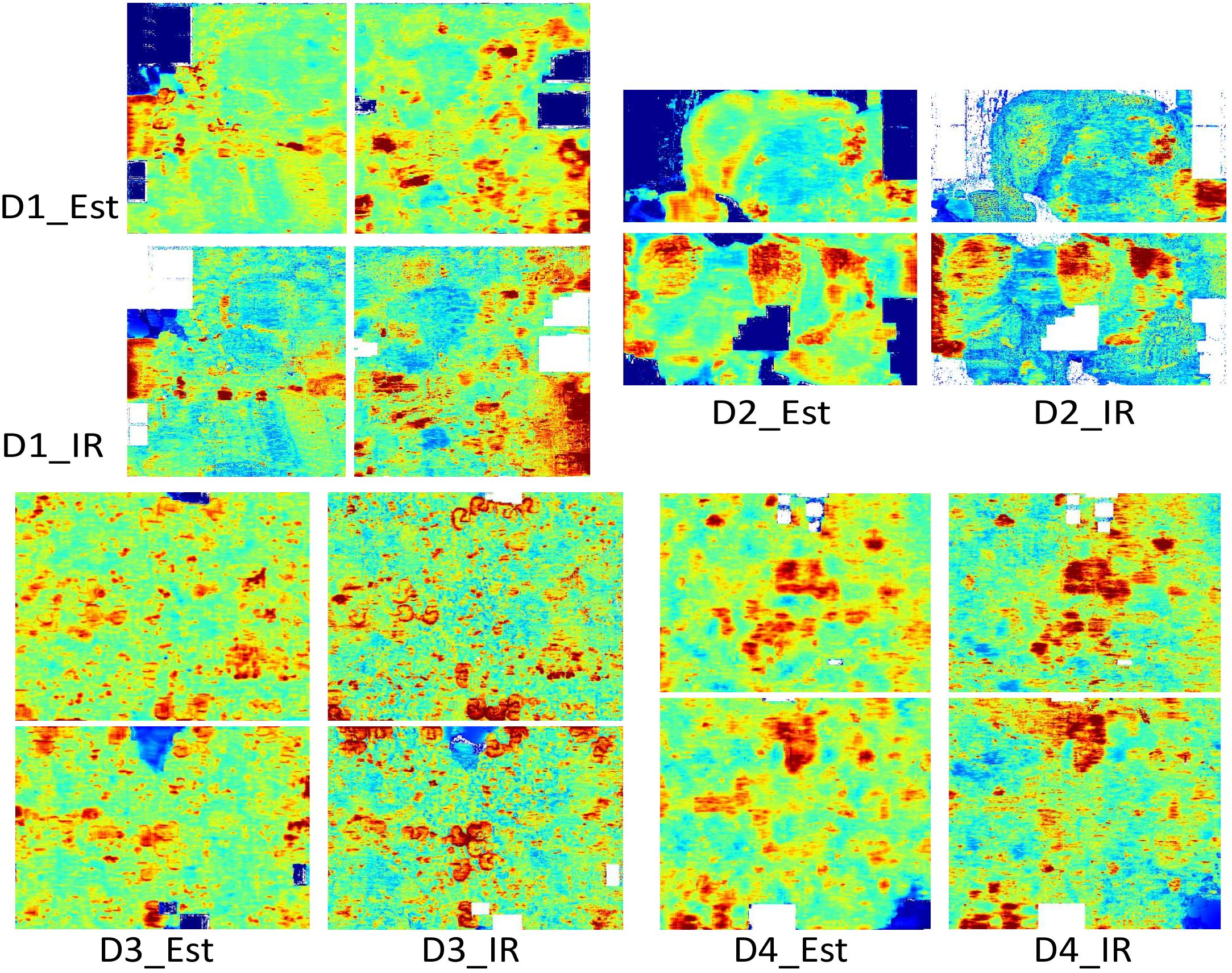}
  \caption{Visualization of estimation and ground truth.}
  \label{visual-min}
  \end{minipage}
\end{figure}

Figure \ref{visual-min} shows visualizations of both ground truth and the prediction results from PowerNet. Only subsets of each design containing IR drop hotspot regions are displayed. Red color indicates higher values while blue corresponds to lower values and white means zero values. The white blocks in ground truth are usually the regions without any cells placed. The comparison shows that PowerNet can capture most IR drop hotspots.

\subsection{Results Measured in Error and Ranking}

 \begin{figure}[!b]
  \centering
  \begin{minipage}[t]{3.5in}
    \includegraphics[width=3.5in]{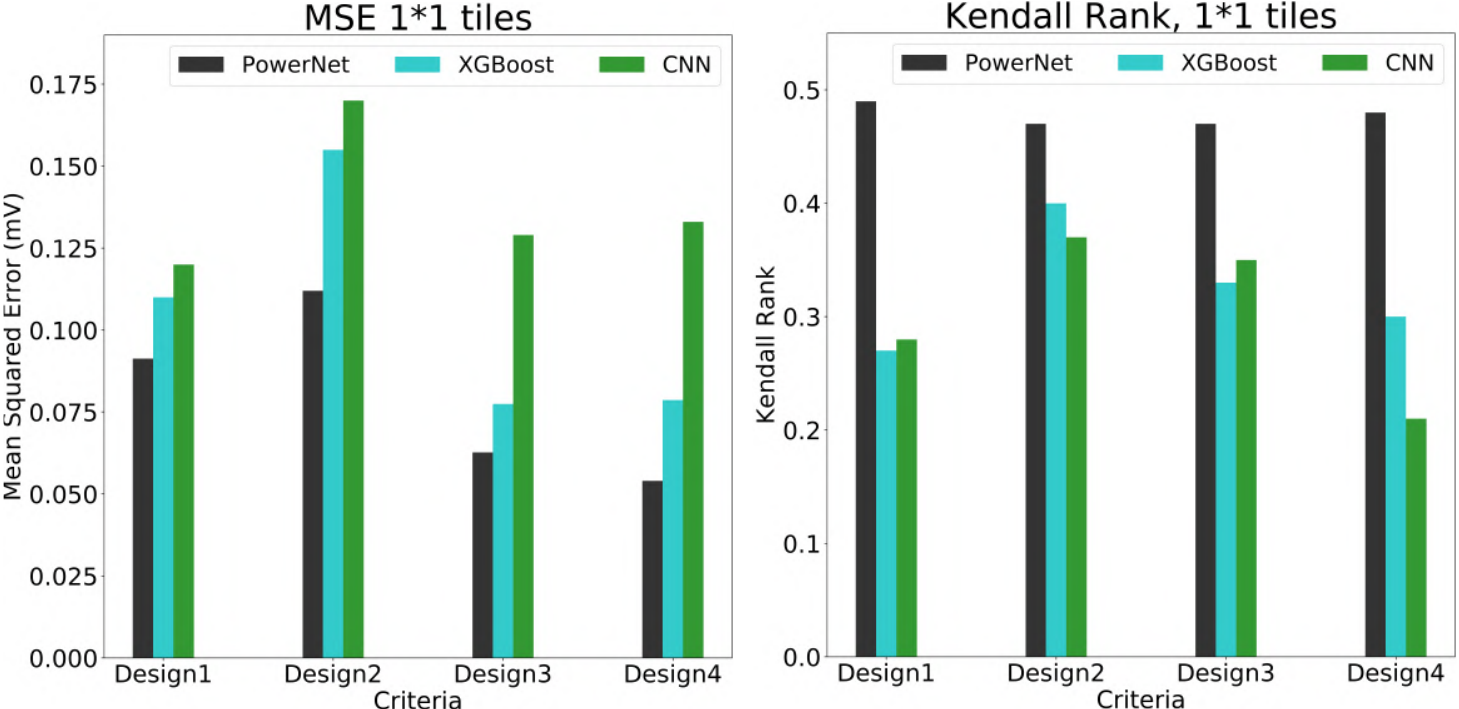}
  \caption{MSE and Kendall ranking coefficient on tiles by IR drop.}
  \label{kendall}
  \end{minipage}
\end{figure}


\fixme{Besides ROC curves, which reflect how well models recognize hotspots, we also measured how models fit and rank tiles according to their IR drop values in Figure~\ref{kendall}. The metrics are mean squared error (MSE) and Kendall rank coefficient~\cite{kendall1948rank} $\tau \in [-1, 1]$ between the estimation and ground-truth IR values for all tiles. A higher value of $\tau$ implies a more accurate ranking of tiles based on IR drop. The MSE and rank coefficients of PowerNet are consistently better than those of other ML methods. Note that a high MSE may be largely contributed by a consistent bias for all inferenced tiles, which means the model always gives a higher or lower prediction for all tiles in one design. In this case, it can still identify those higher-IR tiles or most serious hotspots if it ranks the IR value of tiles accurately.}

\subsection{Inference Time Comparison}
\begin{table}[!h]
  \centering
  \vspace{-2mm}
  \caption{Inference Time Comparison.}
  \vspace{-1mm}
  \label{tbl:time}
  \resizebox{\linewidth}{!}{%
  \begin{tabular}{c | c c c c c}
 	\hline
Method     &  Commercial Tool   &  PowerNet  &  CNN  & XGBoost \\
	\hline
    \multirow{2}{*}{Time} & \multirow{2}{*}{2.5 hour}  &  \multirow{2}{*}{5 min} &  \multirow{2}{*}{1.5 min}  &  \multirow{2}{*}{1.5 min} \\
              &         &      &            &      &         \\
	\hline
  \end{tabular}
  }
\end{table}

The runtimes of the commercial IR drop analysis tool and ML inferences are measured on a design with around two million cells. Results are shown in Table \ref{tbl:time}. PowerNet achieves a $30\times$ speedup over the commercial tool. \fixme{For a fair comparison, the 2.5 hour for the commercial tool only includes analysis time. Its overall runtime is more than 4 hours.} Other ML methods are even faster than PowerNet, but are less accurate. PowerNet is slower than the baseline ML methods because its CNN $f$ generates $N$ outputs $o_j$ for each tile.

\subsection{IR Drop Mitigation in Design Flow}
We also integrated PowerNet into a design flow to mitigate the IR drop of MD1 and MD2. Based on PowerNet's estimation, we enhanced the local power grid (PG) in hotspot regions. 
Notice that the hotspot portions of MD1 and MD2 are much lower than D1 to D4 in the training set. This is because MD1 and MD2 were already close to tapeout and most serious IR drop problems were already fixed, making further IR drop mitigation even more challenging.

Table \ref{tbl:t_ir_mit} shows the IR drop mitigation result. We only add very thin PG straps (0.04 \si{\micro\meter}) at the \fixme{PowerNet-estimated} hotspots. This is the simplest and most basic fixing method. We choose such conservative fixing method to prevent occupying too many routing resources. ``All IR'' and ``Hotspot IR'' mean the averaged IR drop values among all tiles and all hotspots. After PG enhancement, the averaged IR drop for all tiles improves only 0.4 \si{\milli\volt}, which indicates that the modification on PG is very small. In comparison, when measured only on hotspots, IR drop improves 4.3 \si{\milli\volt} and 2.6 \si{\milli\volt}. It shows that PG enhancement is effective at the right places. With such a limited amount of modification in PG, 23\% of IR drop violation cells or around 30\% of hotspots are mitigated. 

\begin{table}[!thb]
  \centering
  \caption{Performance on IR Drop Mitigation}
  \label{tbl:t_ir_mit}
  \resizebox{0.95\linewidth}{!}{%
  \begin{tabular}{l | c c c c}
 	\hline
	\multirow{2}{*}{Design MD1} & Violated  & \#  & All     & Hotspot \\
	                            &  Cell     &    Hotspots   & IR (mV)  & IR (mV)\\
	\hline
	Before Mitigate  &  22185  &  5092  &   26.4   &  66.6   \\
	After  Mitigate  &  17052  &  3778  &   26.0   &  62.3   \\
	Improvement      &    23\% &   26\% &   0.4    &   4.3   \\
	\hline
	\multirow{2}{*}{Design MD2}  & Violated  & \#  & All    & Hotspot \\
	                             &  Cell     &    Hotspots   & IR (mV) & IR (mV)  \\
	\hline
	Before Mitigate &  31097  &  3627 &    31.4  &   62.2   \\
	After  Mitigate &  23941  &  2489 &    31.0  &   59.6   \\
	Improvement     &    23\% &  31\% &    0.4   &    2.6   \\
	\hline
  \end{tabular}
  }
    \vspace{-2mm}
\end{table}
\section{Discussion}

\subsection{PowerNet vs. Previous ML Models}

\begin{figure}[!b]
  \centering
  \begin{minipage}[t]{3.3in}
    \includegraphics[width=3.3in]{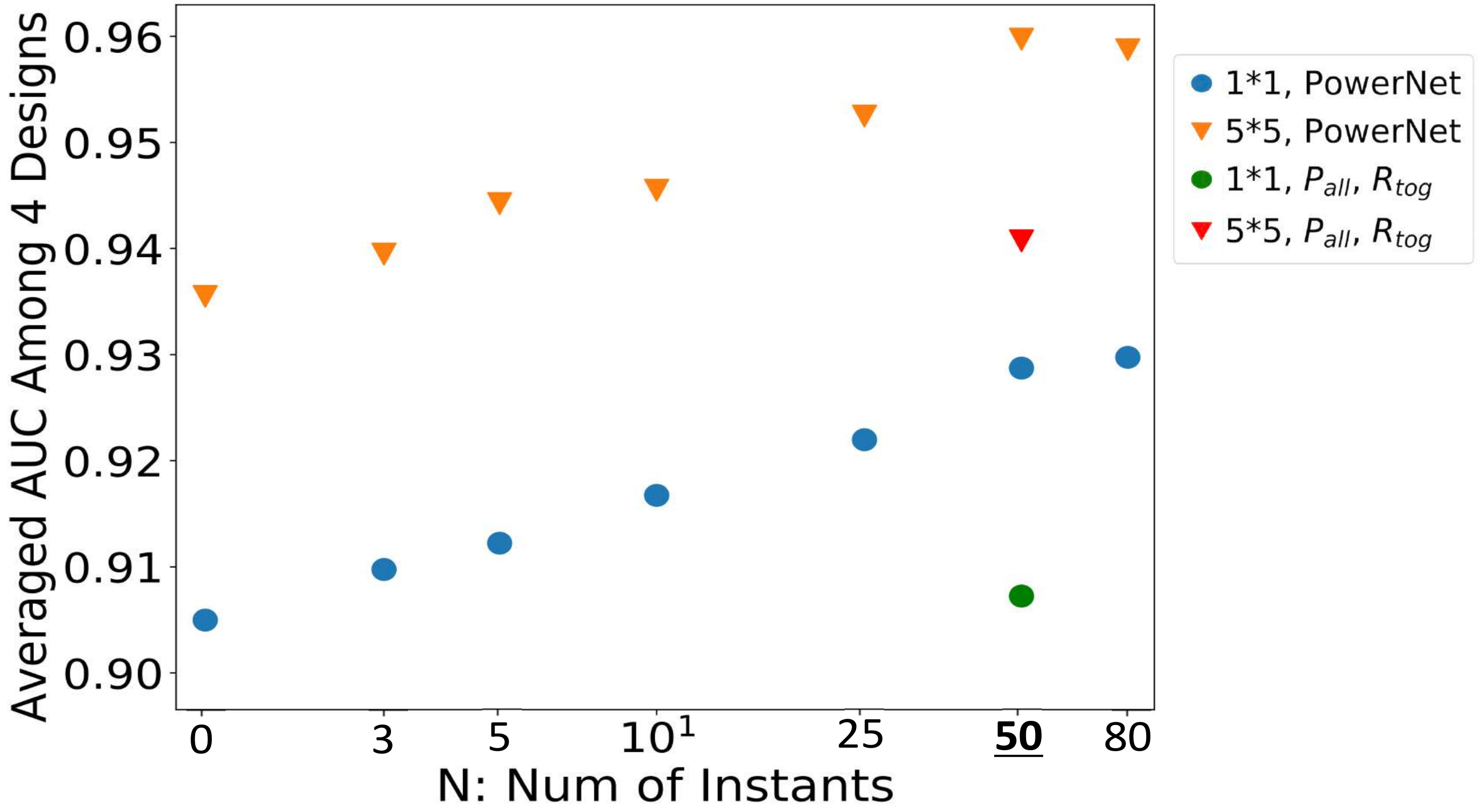}
  \caption{Effect of number of instants $N$ on performance.}
  \label{num_inst}
  \end{minipage}
  \vspace{-1mm}
\end{figure}

We highlight four weaknesses of previous CNN and XGBoost baseline models that prevent them from outperforming PowerNet. First, unnecessary features can confuse ML models. If cell coordinates and time information are used as features but do not directly correlate with IR drop, a model can overfit to designs in the training set. Other features such as cell capacitance can be redundant when power is already provided. To verify this, we implemented an XGBoost model without cell coordinates or time information, and its averaged $1\times 1$ tiles AUC improved from 0.84 to 0.865. When we further removed cell capacitance from features, the averaged AUC remained at 0.865.
Second, different feature formats make the model inefficient. Notice that cell features are one dimensional but map features are two dimensional. For XGBoost, map features must be converted into one dimension, which loses spatial information. For CNN, cell features must be provided through a fully connected layer. In such an unusual CNN structure, cell features tend to be overwhelmed by more than 10,000 outputs from the 50 channels in the last convolutional layer. \fixme{In comparison, PowerNet only uses two-dimensional features.} 
Third, power information may not be fully utilized. When only overall power $P_{all}$ is chosen as a feature, the rich information from other power types $P_i, P_s, P_{sca}$ is lost. Advanced ML models like CNN are complex enough to learn patterns from different power types.
Fourth, time information is not well incorporated or captured. Baselines do not have features like the time-decomposed power maps in PowerNet to measure the worst transient local IR drop.

Figure \ref{num_inst} isolates the contribution of including both time decomposition and multiple power types in variations of PowerNet. 
Average inference AUC accuracy over D1 to D4 is plotted on the Y-axis and the X-axis shows the number of sampled time instants. A higher $N$ means sampling more time instants and generating more corresponding power maps $P_t[\,j]$ within a given clock period $T$. For any region, more power maps better approximate its actual transient power.
The ``N=0'' indicates no time-decomposed power is adopted at all.
As expected, the time-decomposed power maps improve accuracy by capturing transient IR drop. When $N>50$, the improvement in accuracy by increasing $N$ diminishes.
Baseline models also differ from PowerNet by only using maps of features ($P_{all}$ and $r_{tog}$) instead of $\{P_{i}$, $P_s$, $P_{sca}$, $P_{all}\}$.  This variation is indicated as red and green marks in Figure \ref{num_inst}, where time-decomposed power maps $P_t[\,j]$ are kept the same for both variations. In addition to the $2.5\%$ accuracy improvement from time decomposition, adopting multiple types of power improves accuracy by more than $2\%$.

\subsection{Time Decomposition Mechanism}
\begin{figure}[!t]
  \centering
  \begin{minipage}[t]{3.5in}
    \includegraphics[width=3.5in]{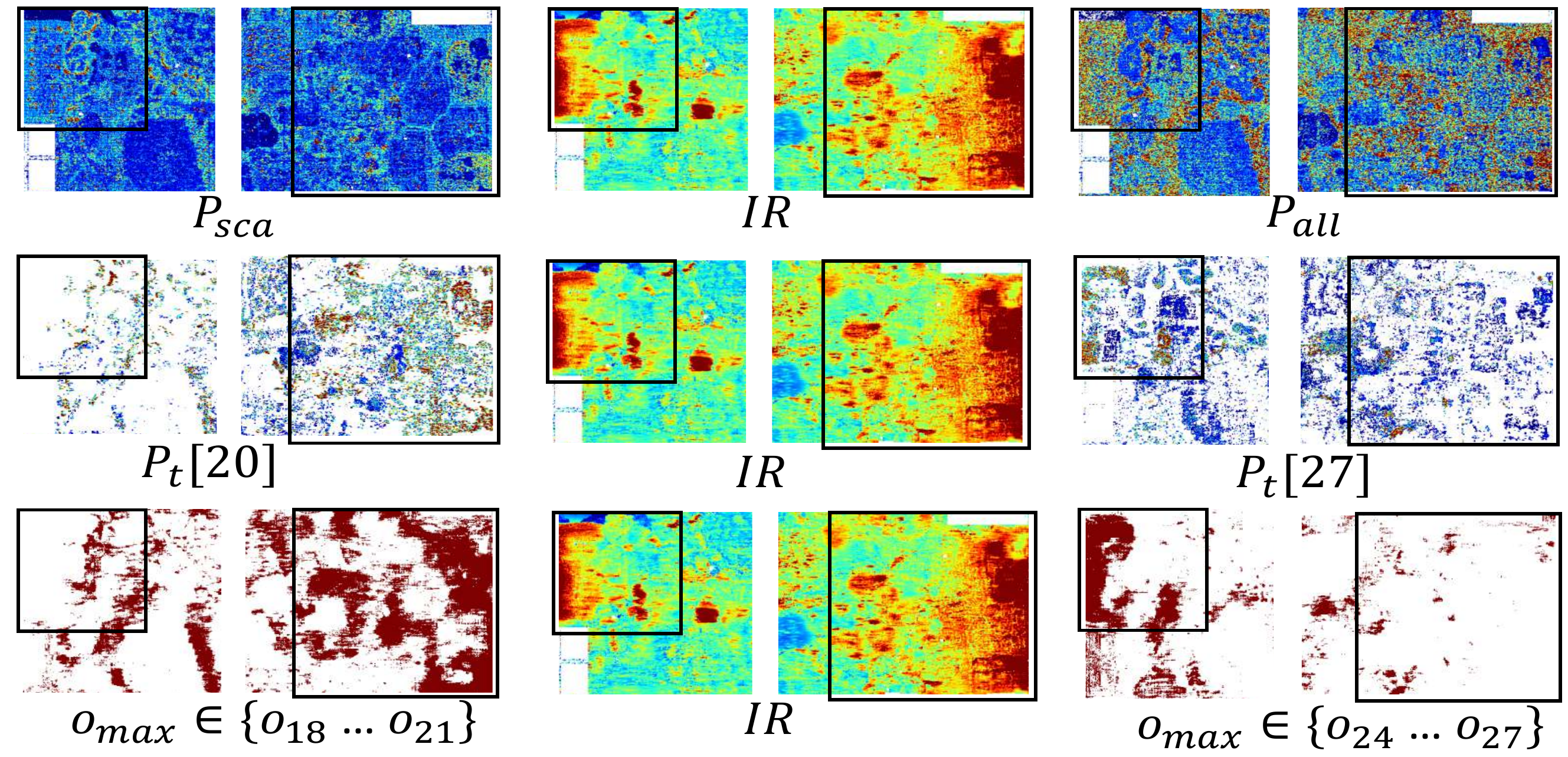}
    \vspace{-3mm}
  \caption{IR drop, power maps and maximum instant distribution of two regions from D1. \fixme{Instants number $N=50$.}}
  \vspace{1mm}
  \label{features}
    \includegraphics[width=3.5in]{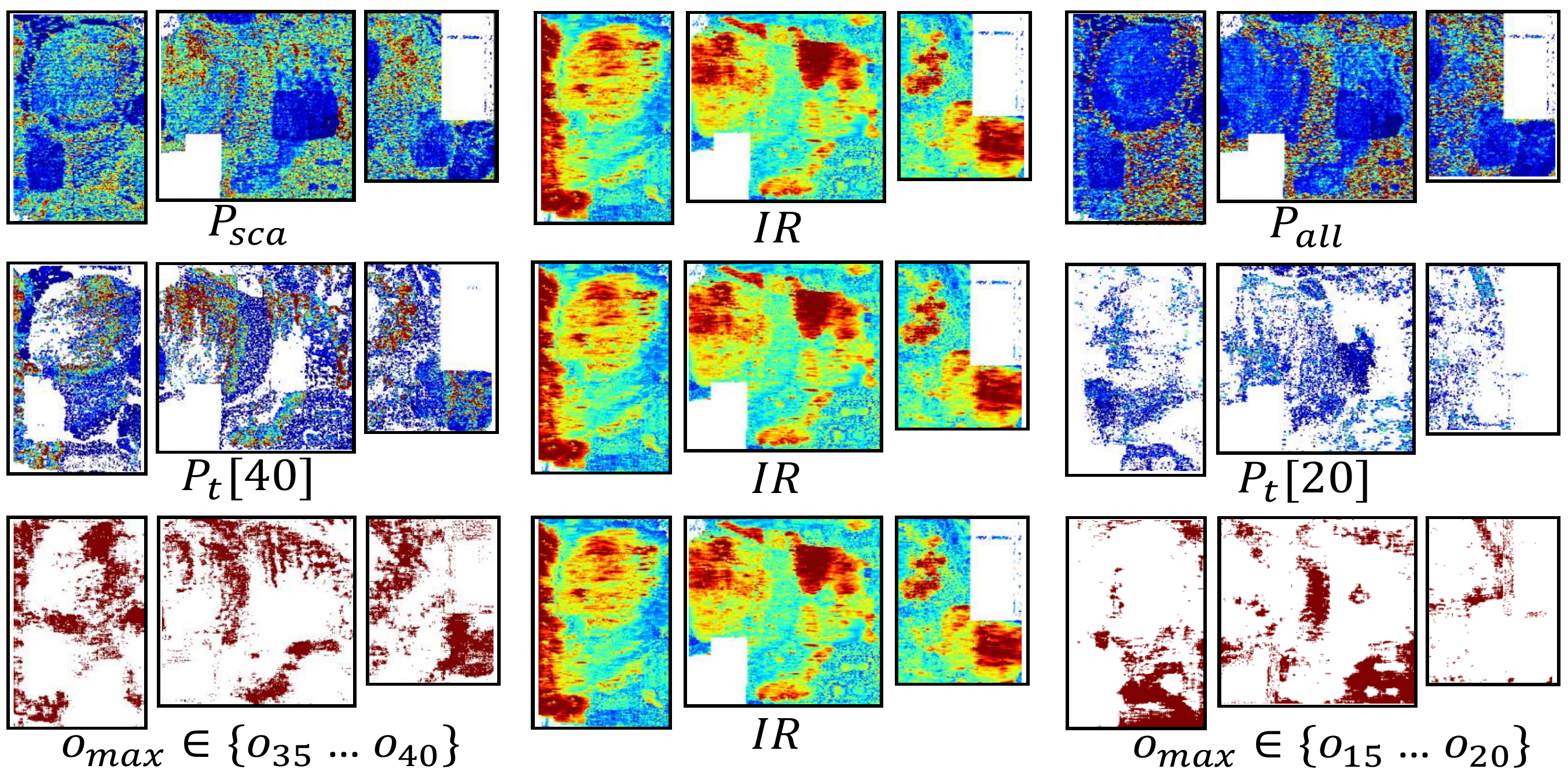}
    \vspace{-3mm}
  \caption{IR drop, power maps and maximum instant distribution of three regions from D2. \fixme{Instants number $N=50$.}}
  \label{features2}
  \end{minipage}
\end{figure}

Figures \ref{features} and \ref{features2} show how the combination of space decomposition and time decomposition helps to explore the potential correlation between power maps and IR drop. It presents the visualization of $IR$, different power maps and maximum instant distribution for the local regions from both D1 and D2. Maximum instant refers to the time instant $j$ selected by maximum structure ($o_j = o_{max}$). The areas of interest are highlighted by black squares. They all contain strong IR drop hotspots. For both D1 and D2, it's difficult to observe much correlation in hotspots between their $\{P_{all}$, $P_{sca}\}$ and $IR$. That indicates training models without any time-decomposed power maps $P_t[\,j]$ can be difficult.

However, the correlation becomes much more clear when power maps $P_t[\,j]$ are provided. In D2, $P_t[40]$ and $IR$ share many common hotspots patterns in highlighted areas. In this case, $o_{40} = f(\{P_{i}$, $P_{s}$ $P_{sca}$, $P_{all}$, $P_{t}[40] \})$ is likely to accurately predict these common hotspot regions. However, another power map, $P_t[20]$, does not share as much hotspot patterns with $IR$. Its output $o_{20}$ may be less accurate. Considering that $P_t[20]$ is much weaker than $P_t[40]$ for most tiles in hotspot regions, we can reasonably assume $o_{40} > o_{20}$, or even $o_{max} = o_{40}$. This is verified by the maximum instant distribution. For every tile, we checked which instant is selected by the maximum structure. For almost all tiles at hotspot regions in D2, their $o_{max} \in \{o_{35}... o_{40}\}$. $P_t[40]$ indeed contributes more information than $P_t[20]$. It is the contribution of the more accurate $P_t[40]$ instead of $P_t[20]$ at these hotspot regions that finally gets captured by the maximum structure. 

Similarly, for D1, the highlighted region on the right correlates well with $P_t[20]$, and region on the left correlates with $P_t[27]$. 
Maximum instant distribution shows that $o_{max} \in \{o_{18}... o_{21}\}$ for most grids in the right region and $o_{max} \in \{o_{24}... o_{27}\}$ for most tiles in the left region.
Then the maximum structure will take $o_t[20]$ for grids on the right and $o_t[27]$ for tiles on the left. In this way, the hotspots caused by transient power at different instants can all be captured.

\subsection{Training Accuracy}

Table \ref{tbl:train} shows the training accuracy for three ML methods. XGBoost shows a higher training accuracy than the CNN baseline, consistent with its better design-dependent performance in previous work~\cite{fang2018machine}. 
PowerNet provides the highest training and inference accuracy among all ML models.

\begin{table}[!t]
  \centering
  \caption{Training Accuracy in ROC AUC (0.01*)}
  \label{tbl:train}
  \resizebox{0.95\linewidth}{!}{
  \begin{tabular}{l | c c c c | c c c c}
 	\hline
 	\multirow{2}{*}{ML Methods}   &   \multicolumn{4}{c|}{$1\times1$ tiles} &                                        \multicolumn{4}{c}{$5\times5$ tiles}    \\
	                  & D1   &  D2 &  D3 &  D4 & D1   &  D2 &  D3 &  D4 \\  
	\hline
    XGBoost           &  93  &  94 &  94 &  94 &  97 &  98 &  98 & 97  \\
	CNN               &  87  &  79 &  80 &  85 &  92 &  85 &  83 & 90  \\
	PowerNet          &  94  &  98 &  95 &  94 &  98 &  99 &  98 & 98   \\ 
	\hline
  \end{tabular}
  }
\end{table}

\subsection{Influence of Resistance}
\label{resistance}

\begin{table}[!b]
  \centering
  \caption{Inference Accuracy in ROC AUC (0.01*)}
  \label{tbl:resist}
  \resizebox{\linewidth}{!}{
  \begin{tabular}{l | c c c c | c}
 	\hline
     ML Methods  & D1   &  D2 &  D3 &  D4 & Ave  \\        
	\hline
	PowerNet ($1\times1$ tiles)  &  92.1  &  95.4 &  91.4 &  92.6 & 92.9  \\ 
	\fixme{PRNet}\;\;\;\;\;($1\times1$ tiles)  &  92.4  &  95.5 &  90.5 &  93.6 & 93.0  \\
	\hline
	PowerNet ($5\times5$ tiles)  &  95.4  &  96.7 &  94.8 &  97.0 & 96.0  \\ 
	\fixme{PRNet}\;\;\;\;\;($5\times5$ tiles)  &  95.7  &  96.8 &  93.2 &  97.5 & 95.8  \\
	\hline
  \end{tabular}
  }
\end{table}

We measured the distribution of resistance in our benchmark design. \fixme{Take D1 for example, the standard deviation in resistance across the whole design is only 2.8\si{\ohm}, $0.6\%$ of its average resistance}. For such a uniform distribution, we chose not to spend extra time calculating per-cell resistance.  However, we did implement another variation of PowerNet where each cell's power is scaled with resistance, denoted as \fixme{``PRNet''} in Table \ref{tbl:resist}. ``Ave'' means accuracy averaged over all four designs. On average, the resistance-scaled PowerNet shows similar accuracy to the original one. This demonstrates that using per-cell resistance as a feature is not necessary for designs with uniform PDNs. By scaling power with resistance, \fixme{``PRNet''} can be further applied to designs with non-uniform PDNs.

\subsection{Vector-Based IR Drop Estimation}
\label{vector}

\begin{figure}[!t]
  \centering
  \begin{minipage}[t]{3.2in}
    \includegraphics[width=3.2in]{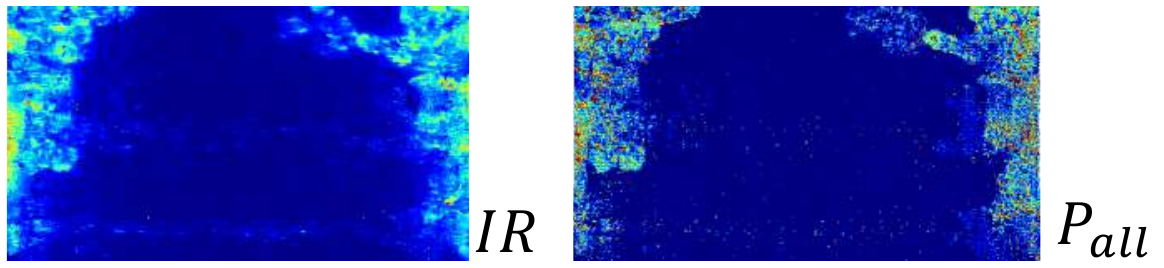}
  \caption{Power and IR drop of vector-based estimation.}
  \label{vector-min}
  \end{minipage}
\end{figure}

We also measured PowerNet's performance on vector-based IR drop. 
\fixme{The PowerNet architecture remains exactly the same, but cell power and IR drop are now} collected when the commercial tool simulates IR drop with given simulation patterns.
Figure \ref{vector-min} shows 
\fixme{the vector-based} power map $P_{all}$ and label $IR$. Unlike the vectorless case in Figure \ref{features} or \ref{features2}, the power of a portion of activated cells is significantly higher than the others. As we mentioned, the correlation between power and IR drop value turns out to be very strong, which largely reduced the estimation difficulty.

We perform vector-based estimation on four other industrial designs VD1 to VD4. All \fixme{models and} procedures are the same as the vectorless case, except using cell power and IR drop from vector-based simulation. Table \ref{tbl:vector} shows vector-based estimation accuracy. As expected, all methods provide better estimation than vectorless scenario. But PowerNet still gives the best accuracy for every single design. The $1\%$ to $2\%$ improvement should not be underestimated when accuracy is already very high. To a certain extent, boosting accuracy from $98\%$ to $99\%$ means reducing half of the errors.

\begin{table}[!h]
  \centering
  \caption{Vector-Based Inference in ROC AUC (0.01*)}
  \label{tbl:vector}
  \resizebox{\linewidth}{!}{
  \begin{tabular}{l | c c c c | c c c c}
 	\hline
 	\multirow{2}{*}{ML Methods}   &   \multicolumn{4}{c|}{$1\times1$ tiles} &                                        \multicolumn{4}{c}{$5\times5$ tiles}    \\
	                  &  VD1  &  VD2 &  VD3 &  VD4 &   VD1 &  VD2 &   VD3 &  VD4  \\  
	\hline
	XGBoost           &  97  &  98 &  98 &  96 &   99 &  97 &   98 & 97 \\
	CNN               &  96  &  93 &  95 &  95 &   98 &  92 &   97 & 96  \\
	PowerNet          &  98  &  98 &  99 &  97 &  100 &  98 &  100 & 98  \\ 
	\hline
  \end{tabular}
  }
\end{table}

\section{Conclusion}
In this paper, we present a CNN-based dynamic IR drop estimator. Unlike existing ML works, our model is general and transferable to new designs. 
We validate the high accuracy of our approach on multiple industrial designs. It takes an order of magnitude less estimation time than commercial tools and significantly outperforms state-of-the-art ML methods in both vector-based and vectorless IR drop scenarios. 
The IR drop mitigation tool guided by our model reduces IR drop by more than 20\% with very limited PG modification.


\section*{Acknowledgments}
This work is partially supported by Semiconductor Research Corporation Tasks 2810.021 and 2810.022 through UT Dallas’ Texas Analog Center of Excellence (TxACE).

\bibliographystyle{IEEEtran}
\begin{spacing}{0.9}
\bibliography{IR_zhiyao.bib}
\end{spacing}
\end{document}